\documentclass[]{article}
\usepackage{arxiv}
\usepackage{lmodern}
\usepackage{tabularx}
\usepackage{array}
\usepackage{abstract}
\usepackage{geometry}
\usepackage{multirow}
\usepackage{boldline}
\usepackage{amssymb,amsmath}
\usepackage{authblk}
\usepackage{ifxetex,ifluatex}
\usepackage[labelfont=bf,justification=centering]{caption}
\usepackage[ruled,linesnumbered]{algorithm2e}
\ifnum 0\ifxetex 1\fi\ifluatex 1\fi=0 
  \usepackage[T1]{fontenc}
  \usepackage[utf8]{inputenc}
  \usepackage{textcomp} 
\else 
  \usepackage{unicode-math}
  \defaultfontfeatures{Scale=MatchLowercase}
  \defaultfontfeatures[\rmfamily]{Ligatures=TeX,Scale=1}
\fi
\IfFileExists{upquote.sty}{\usepackage{upquote}}{}
\IfFileExists{microtype.sty}{
  \usepackage[]{microtype}
  \UseMicrotypeSet[protrusion]{basicmath} 
}{}
\makeatletter
\@ifundefined{KOMAClassName}{
  \IfFileExists{parskip.sty}{%
    \usepackage{parskip}
  }{
    \setlength{\parindent}{0pt}
    \setlength{\parskip}{6pt plus 2pt minus 1pt}}
}{
  \KOMAoptions{parskip=half}}
\makeatother
\usepackage{xcolor}
\IfFileExists{xurl.sty}{\usepackage{xurl}}{} 
\IfFileExists{bookmark.sty}{\usepackage{bookmark}}{\usepackage{hyperref}}
\hypersetup{
  hidelinks,
  pdfcreator={LaTeX via pandoc}}
\usepackage{longtable,booktabs}
\usepackage{etoolbox}
\makeatletter
\patchcmd\longtable{\par}{\if@noskipsec\mbox{}\fi\par}{}{}
\makeatother
\IfFileExists{footnotehyper.sty}{\usepackage{footnotehyper}}{\usepackage{footnote}}
\makesavenoteenv{longtable}
\usepackage{graphicx}
\makeatletter
\def\maxwidth{\ifdim\Gin@nat@width>\linewidth\linewidth\else\Gin@nat@width\fi}
\def\maxheight{\ifdim\Gin@nat@height>\textheight\textheight\else\Gin@nat@height\fi}
\makeatother
\setkeys{Gin}{width=\maxwidth,height=\maxheight,keepaspectratio}
\makeatletter
\def\fps@figure{htbp}
\makeatother
\title{\textbf{SkIn: Skimming-Intensive Long-Text Classification Using BERT
for Medical Corpus}}
\author[1]{Yufeng Zhao$^1$, Haiying Che$^{*}$}
\affil[1]{Beijing Institute of Technology, Beijing, China}

\date{}

\begin{document} 
\maketitle
\vspace{-3em}
\pagenumbering{arabic}
\begin{abstract}
BERT is a widely used pre-trained model in natural language processing. 
However, since BERT is quadratic to the text length, the BERT model is difficult to be used directly on the long-text corpus. 
In some fields, the collected text data may be quite long, such as in the health care field. Therefore, to apply the pre-trained
language knowledge of BERT to long text, in this paper, imitating the skimming-intensive
reading method used by humans when reading a long paragraph, the \textbf{Sk}imming-\textbf{In}tensive
Model (SkIn) is proposed. It can dynamically select the critical information in the text so that
the sentence input into the BERT-Base model is significantly shortened, which can effectively
save the cost of the classification algorithm. Experiments show that the SkIn method has
achieved superior accuracy than the baselines on long-text classification datasets in the medical
field, while its time and space requirements increase linearly with the text length, alleviating
the time and space overflow problem of basic BERT on long-text data. 
\\
\end{abstract}

\section{1. Introduction}

A large number of comment texts are generated  on social media every time, and in some fields, some of these texts have long
lengths. Especially in the medical field\cite{tri_fea, ref3, dataset2}, the text describing the
disease, the treatment effect, etc., may have a considerable length.
Mining the opinions and sentiments contained in these comments helps  
track people's attitudes and behaviors\cite{App1}, look for similar patients and provide advice to new patients\cite{App2, App3}, 
improve the survivorship of patients\cite{App4}, identify potential disease hazards\cite{App5}, 
collect statistical data to help scientific research in the medical field\cite{tri_fea},
and prevent potential drug adverse effects and medical accidents\cite{App2, App8}.
However, manual classification of such massive data will cost a lot\cite{App9}, so we need to find an automatical way.
However, because of the length of these texts, such tasks are quite challenging.

In the \textbf{N}atural \textbf{L}anguage \textbf{P}rocessing (NLP)
field, pre-trained models can introduce a priori language knowledge, accelerating and optimizing the model training.
BERT\cite{ref1} performs to some extent among the pre-trained language models, so it is widely used in NLP tasks\cite{bertapp1}. However,
BERT is based on Transformer structure\cite{ref2}, which
requires the time and space to the level of quadratic power of the input length for the intermediate results. This problem poses a
challenge for applying the BERT model in long-text corpus.

When reading a long text, humans sometimes skim the full text
quickly and then pay more attention to a critical segment for intensive reading. Therefore,
with the inspiration of such skimming-intensive reading skill, we propose the \textbf{Sk}imming-\textbf{In}tensive method (SkIn). We
first use a classification algorithm with lower accuracy but a cheap cost
to classify roughly, and then we select the most crucial segment through the classification process. Finally, we input the
selected segment into a more precise classification model. In order
to select the most crucial segment of a sentence, we propose the
self-adaptive attention method to judge the importance of each segment.

We apply this method to \textbf{C}orpora for \textbf{M}edical \textbf{S}entiment(CMS)\cite{ref3}, a long-text
classification dataset in the medical field. Compared with the basic BERT
model, SkIn achieves a improvement of accuracy while significantly
reducing the cost of time and space. The increase of the cost of SkIn is much more slower than the basic BERT model. It saves up to more than 90\% of the space
cost compared with the baselines.

\section{2. Background}

\textbf{BERT pre-trained language model}\cite{ref1}, which
stands for \textbf{B}idirectional \textbf{E}ncoder
\textbf{R}epresentations from \textbf{T}ransformers, is based on the
Transformer encoder architecture. It embeds tokens into word
vectors and then uses the multi-head self-attention layers to 
integrate the relevant information in the whole sentence, for which some scholars
call it "contextual embedding"\cite{tri_fea, ref4}. Due to space
limitation of this paper, please refer to reference\cite{ref2} for a detailed description of the multi-head self-attention and Transformer. We can briefly describe a single
self-attention mechanism layer applied in BERT by Eq.(1), where
$K \in \mathbb{R}^{L \times d_{k}}$, $L$ is the length of the input and
$d_{k}$ is the embedding dimension. If the number of the attention
layers in BERT is recorded as $N$, BERT has
$O(Nd_{k}L^{2})$ time \& space cost. The BERT-Base model, the most versatile and popular one among BERT, has an embedding dimension
of 768 and 12 layers of attention\cite{ref1}, which makes the
GPU memory overflow easily during the training when $L$ is large. Even though reducing the
batch size can save the GPU memory, the training time with reduced batches will be incredibly prolonged. Therefore, BERT sets an input length
limit of 512 tokens to avoid the overflow problem, but it does not solve the problem essentially.

\begin{equation}
SelfAttention(K) = softmax( \frac{KK^{T}}{\sqrt{d_{k}}})K
\end{equation}

Once BERT was proposed, it broke the best scores on several NLP tasks, and its pre-training can introduce rich language knowledge into our next processing model.
Because of this advantage, we still try to find ways to apply it despite the difficulty of lengthy text.
Google\footnote{https://github.com/google-research/bert} provides
open-source pre-trained BERT models with different complexities. For example, the
BERT-Tiny model, a concise version of BERT-Base, has an embedding dimension of 128 and 2 layers of
attention. The BERT pre-trained models used in this paper are all
provided by Google.

\textbf{Related works}. Some scholars solved this problem by truncating the input in fixed ways\cite{howto, ref5}.
Chen et al.\cite{ref5} extracted two sentences shorter than a certain length, the one
from the beginning, and the other from the end of the text. Then
they input them into the BERT-Base model for
classification. Nevertheless, this kind of method suffers from a loss of information that could help understand the sentence\cite{TextGuide}.

Another method is segmenting the text into many short segments, then
using a downstream network to establish semantic associations between different segments\cite{howto, tri_fea, ref6}. Pappagari et al.\cite{ref6}
proposed a slide-window model, where the
text is segmented into fixed length blocks and input into BERT.
Then, an LSTM\cite{ref7} network or Transformer is used to fuse the output of different blocks into a global
feature. This method seems to avoid the information loss of the truncation method, but experiments show that it has no significant improvement in accuracy, 
and still has a quadratic memory cost (see \S \nameref{seg53}).

Some scholars\cite{ot1, ot2, ot3} choose to optimize the Transformer.
Beltagy et al.\cite{ot2} used local windowed attention for shortening the length of sentences to be processed simultaneously by attention.
By such methodology, the cost of the Transformer can be saved, while the upper length limit of BERT can be increased.
However, such an increment is often limited and accompanied by the loss of information that may contain essential information.

Min et al.\cite{select1} proved that there is a critical segment in almost all
long text by human reading experiment, and the information obtained in this segment is similar to the whole text. 
With this idea, they divided a paragraph into sentences and proposed a method to select the critical sentences using the encoder-decoder architecture.
Ding et al.\cite{ref8} proposed a method called MemRecall to give a score to each part to select the key segment. 
Their methods achieved excellent results in several Q\&A and classification datasets.
Fiok et al.\cite{TextGuide} also proposed the Text Guide method to select the key segment.
Selecting the critical segment significantly reduces the algorithm overhead without losing essential information.
This paper also follows the idea of key-segment selection.

\section{3. Methodology \& Model}

\textbf{Problem definition}. Given the input text
$s \in \mathbb{Z}^{L}$, the model is expected to predict the
sentence classification $\widehat{y}$ posteriorly. Where the number of classes is $U$, in other words,
$\widehat{y} \in \{ 1,2,\ldots,U \}$.

\textbf{Methodology.} Inspired by the previous work\cite{ref8}, we also assume that there is a key segment
$s^{+} \subset s$, which can be classified similarly to $s$ by a
certain $Classifier$ as Eq.(2):

\begin{equation}
Classifier( s^{+} ) \approx Classifier(s) \\
\end{equation}

Therefore, how to find this key segment is an essential problem.
An intuitive way is to tag the key segment manually and train a Sequence Labeling\cite{ref9} model to segment the text.
However, such manual tagging is not only tedious but also affected by subjective factors easily.
So to calculate the importance of each segment without additional labeling, we propose the
\textbf{S}elf-\textbf{a}daptive \textbf{A}ttention mechanism (SaA).

\begin{figure}
  \begin{center}
\includegraphics[width=0.70\textwidth]{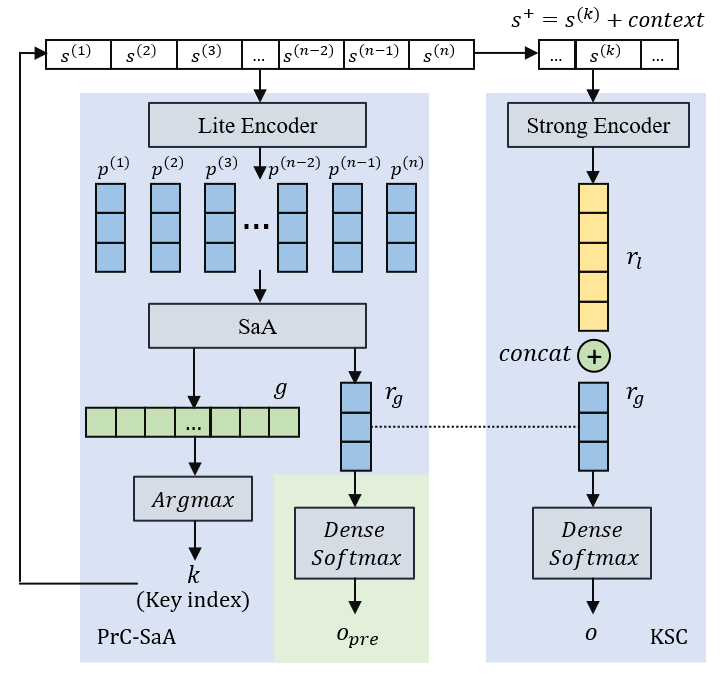}
\caption{The architecture of SkIn model.}\label{fig1}
\end{center}
\end{figure}

\textbf{Model summary}. As shown in Figure \ref{fig1}, we first divide the long
text $s$ into $n$ segments of length $l$, that is
$(s^{(1)},s^{(2)},\ldots,s^{(n)})$. Then, the skimming method, called
\textbf{Pr}evious \textbf{C}lassification and
\textbf{S}elf-\textbf{a}daption \textbf{A}ttention (PrC-SaA) is used to
quickly generate a weight for each segment and a previous classification result related to this weight. The classification result is used for training the PrC-SaA while the score is used for selecting the key segment. 
We select the segment $s^{(k)}$ with the highest score and its context as the key segment $s^{+}$. Then we
input $s^{+}$ into the intensive method called \textbf{K}ey \textbf{S}egment \textbf{C}lassification (KSC) for precise classification. 
The PrC-SaA is to imitate the skimming reading of human, and the KSC is the intensive reading,
for which our model is called \textbf{Sk}imming-\textbf{In}tensive Model (SkIn).

\subsection{3.1 Skimming: Previous Classification and Self-adaption
Attention}

The PrC-SaA consists of a Lite Encoder and a SaA module.

\textbf{Lite Encoder} quickly encodes all the input segments into semantic vectors with a lightweight model. 
It outputs a semantic vector $p^{(i)} \in \mathbb{R}^{d_{g}}$ for each segment $s^{(i)}$ with embedding dimension ${d_{g}}$ as shown in Eq.(3):

\begin{equation}
p^{( i )} = LiteEncoder( s^{( i)})
\end{equation}

In our model, the BERT-Tiny pre-trained model which is much cheaper than the BERT-Base, and a global average
pooling layer are used as the Lite Encoder, as
shown in Eq.(4) and (5), where $v^{(i)} \in \mathbb{R}^{l \times d_{g}}$ is the embedding
matrix generated by BERT-Tiny of $s^{(i)}$.

\begin{equation}
v^{( i )} = BERT_{Tiny}( s^{( i )} ) \\
\end{equation}

\begin{equation}
p^{( i )} = \frac{\sum_{j = 1}^{l}v_{j}^{( i )}}{l}
\end{equation}

\textbf{Self-adaptive Attention} (SaA) gives weight to each segment's encoding
vector $p^{( i )}$ and averages them into a global semantic vector by the given weight as shown in Eq.(7) and Eq.(8). 
Previously, each $p^{(i)}$ from the lite encoder layer is merged into the global encoding matrix
$p \in \mathbb{R}^{n \times d_{g}}$ as shown in Eq.(6).

\begin{equation}
p = \lbrack p^{( 1 )};\ p^{( 2 )};\ldots;\ p^{( n - 1 )};\ p^{( n )} \rbrack
\end{equation}

Then, the self-adaptive attention is used as Eq.(7) and (8). Unlike the
general attention mechanism\cite{ref10}, SaA uses a trainable
model parameter $W_{a} \in \mathbb{R}^{1 \times n}$ as one of the inputs of the attention mechanism. 
So the SaA is more like a dense layer than an attention layer.
$g \in \mathbb{R}^{n}$ holds the weight score given to each segment,
$r_{g} \in \mathbb{R}^{d_{g}}$ is the global semantic vector. Although
the factor $1/\sqrt{d_{g}}$ will reduce the variance of the result of
$softmax$, it can put the intermediate result $W_{a}p^{T}$ in a suitable
range and avoid potential overflow and gradient problems\cite{ref2}.

\begin{equation}
g = softmax( \frac{W_{a}p^{T}}{\sqrt{d_{g}}\ } )
\end{equation}

\begin{equation}
r_{g} = gp
\end{equation}

To train the PrC-SaA, the skimming part, $r_{g}$ is input into a dense layer activated by $softmax$ to generate a
vector $o_{pre} \in \mathbb{R}^{U}$ that contains the probability of the sentence belongs to each category as Eq.(9). Where
$W_{op} \in \mathbb{R}^{U \times d_{g}}$ and
$b_{op} \in \mathbb{R}^{U}\ $are the model parameters.

\begin{equation}
o_{pre} = softmax( W_{op}r_{g} + b_{op} )
\end{equation}

Therefore, we can use $o_{pre}$ to train the PrC-SaA, which can judge the segment with the most significant importance, only through the labels of the
original dataset without manually marking the most important segments.

\subsection{3.2 Key Segment Selection}

After the training of the skimming model, we input
each long text in the dataset into PrC-SaA and get its weight vector
$g$, then select its key segment $s^{+}$ by Algorithm \ref{a1}.

\begin{center}
\begin{minipage}{0.47\textwidth}
\IncMargin{1em} %
\begin{algorithm}[H]

    \caption{Key Segment Selection}\label{a1}
    \SetKwInOut{Input}{\textbf{Input}}\SetKwInOut{Output}{\textbf{Output}} %

    \Input{
        \\
        Weight vector $\mathbf{g}$\;\\
        Long text $\mathbf{s}$\;\\
        Number of segments $\mathbf{n}$\;\\
        Length of each segment $l$\;\\
        }
    \Output{
        \\
        Selected key segment $\mathbf{s}^{\mathbf{+}}$\;\\
        }
    \BlankLine

    Get the key index $\mathbf{k}{=Argmax(}\mathbf{g})$\; %
    \uIf {$\mathbf{k} == 0$} {\emph{return} $\mathbf{s}[0:l+l/2]$}
    \uElseIf {$\mathbf{k} == \mathbf{n}-1$} {\emph{return}
    $\mathbf{s}[l*(\mathbf{n}-1)-l/2:l*\mathbf{n}]$}
    \Else {\emph{return}
    $\mathbf{s}[l*(\mathbf{k}-1)-l/4:l*\mathbf{k}+l/4]$}
\end{algorithm}
\DecMargin{1em}
\end{minipage}
\end{center}

Merging too much context may lead to irrelevant information and excessive overhead, while too little context may lead to losing essential information when key parts cross the edge of the segment.
Therefore, by Algorithm \ref{a1}, the length of each key segment is fixed at $1.5l$, which includes the content of the segment with the highest score.
If the skimming model is correctly set and adequately trained, the selection of key
segment can effectively prevent irrelevant information from entering the
precise intensive model, and as a result, the model performance can be optimized (see \S \nameref{seg52}).

\subsection{3.3 Intensive: Key Segment Classification}

Since the length of the key segment is significantly shorter than
the original text, a heavy classification model with
high accuracy can be used on shortened text. That is the KSC module.

The KSC consists of a Strong Encoder and a vector concatenation operation.

\textbf{Strong Encoder}. Similar to the Lite Encoder, the input key
segment $s^{+}$ is firstly encoded by a high-precision text encoding
layer into a local semantic vector $r_{l} \in \mathbb{R}^{d_{l}}$,
where $d_{l}$ is the embedding dimension of this encoder, as shown in
Eq.(10).

\begin{equation}
r_{l} = StrongEncoder( s^{+} )
\end{equation}

In our model, similar to the PrC-SaA, the BERT-Base pre-trained model and a global average
pooling layer are used as the Strong Encoder, as shown in Eq.(11) and (12).

\begin{equation}
e_{l} = BERT_{Base}( s^{+} )
\end{equation}

\begin{equation}
r_{l} = \frac{\sum_{i = 1}^{1.5l}e_{l,i}}{1.5l}
\end{equation}

\textbf{Vector concatenation}. In order to complete the part of global information, the
local encoding vector $r_{l}$ and the global encoding vector $r_{g}$
generated in the PrC-SaA are concatenated as the final feature
$r \in \mathbb{R}^{d_{g} + d_{l}}$, as shown in Eq.(13).

\begin{equation}
r = \lbrack r_{l},r_{g} \rbrack
\end{equation}

This final feature is output via a dense layer similar with PrC-SaA, as shown in Eq.(14), where $o \in \mathbb{R}^{U}$, whose
content is the probability that the text belongs to each category,
$W_{o} \in \mathbb{R}^{U \times (d_{g} + d_{l})}$ and
$b_{o} \in \mathbb{R}^{U}\ $are the model parameters.

\begin{equation}
o = softmax( W_{o}r_{g} + b_{o} )
\end{equation}

\subsection{3.4 Training Methodology}

\textbf{Stage 1.} The PrC-SaA model is trained in advance, during which the $o_{pre}$ is taken
as the output, and the cross-entropy function characterizes the loss as Eq.(15).

\begin{equation}
Loss_{PrC} = CrossEntropy( o_{pre} )
\end{equation}

\textbf{Stage 2.} After the previous training, for each long text in the
dataset, PrC-SaA is used to predict a weight score, and then
Algorithm 1 extracts its key segment. 

\textbf{Stage 3.} After the whole dataset is segmented and selected, the joint training of the whole SkIn
model is conducted. PrC-SaA still accepts the input of all segments
$(s^{(1)},s^{(2)},\ldots,s^{(n)})$, while KSC accepts the input of key
segment $s^{+}$. Instead of $o_{pre}$, $o$ is used as the
output. The cross-entropy function characterizes the loss as Eq.(16).

\begin{equation}
Loss = CrossEntropy( o ) \\
\end{equation}

\section{4. Experiment}

To verify the effectiveness of SkIn model, we apply it to the CMS
dataset collected by Yadav et al.\cite{ref3}. Moreover, we measure its prediction accuracy and time \& memory cost.

\subsection{4.1 Dataset}

The CMS dataset is composed of two sub-datasets. One is the
\textbf{Medical-Condition} dataset, composed of social media users'
comments on their clinical symptoms.  As shown in Table \ref{table1}, these comments are divided into three categories according to the progress of clinical symptoms:
symptoms \textbf{Exist}, symptoms \textbf{Recover}, and symptoms \textbf{Deteriorate}. The other is the \textbf{Medication} dataset, which is
composed of comments on the effects of medical treatment. As shown in Table \ref{table2}, these comments are divided into three categories according to the effect of medical treatment:
\textbf{Effective} treatment, \textbf{Ineffective} treatment, and \textbf{S}erious
\textbf{A}dverse \textbf{E}ffect (\textbf{SAE}).

\begin{center}
  \begin{table}[!htbp]
    \caption{Examples of \textbf{Medical-Condition} dataset. The underlined are the basis for labeling.}\label{table1}
    \centering
  \includegraphics[width=\textwidth]{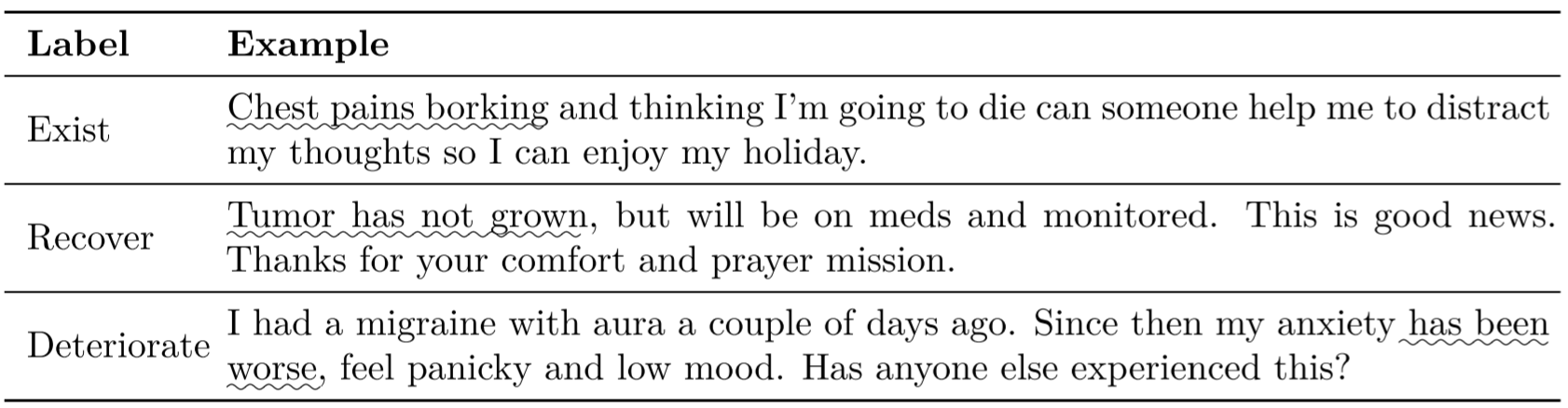}
  \vspace{-1.5em}
  \end{table}
\end{center}

\begin{center}
  \begin{table}[!htbp]
    \caption{Examples of \textbf{Medication} dataset. The underlined are the basis for labeling.}\label{table2}
    \centering

  \includegraphics[width=\textwidth]{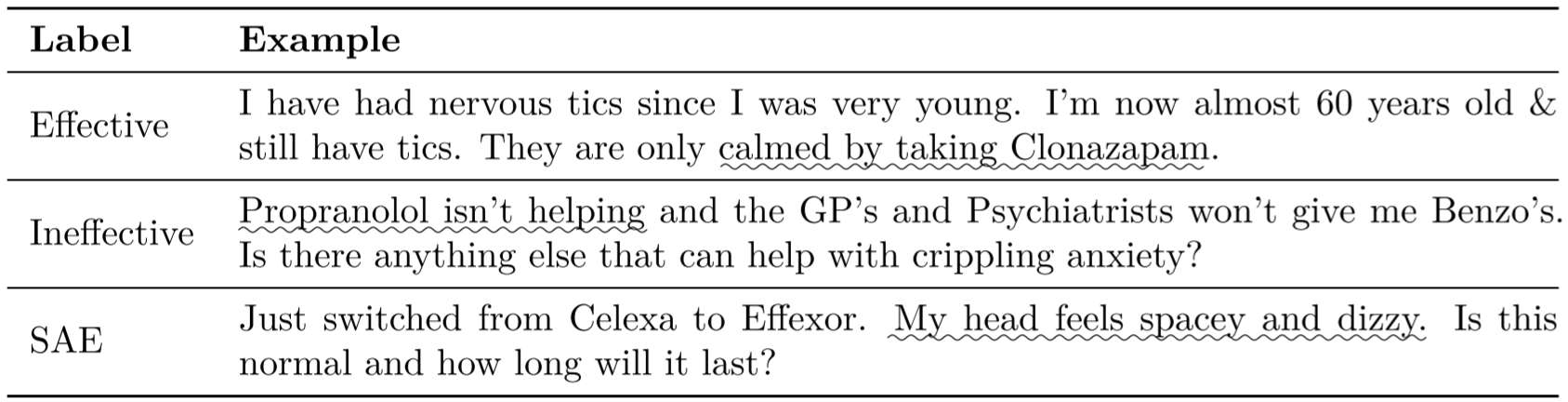}
  \vspace{-1.5em}
\end{table}
\end{center}

\begin{center}
  \includegraphics[width=0.99\textwidth]{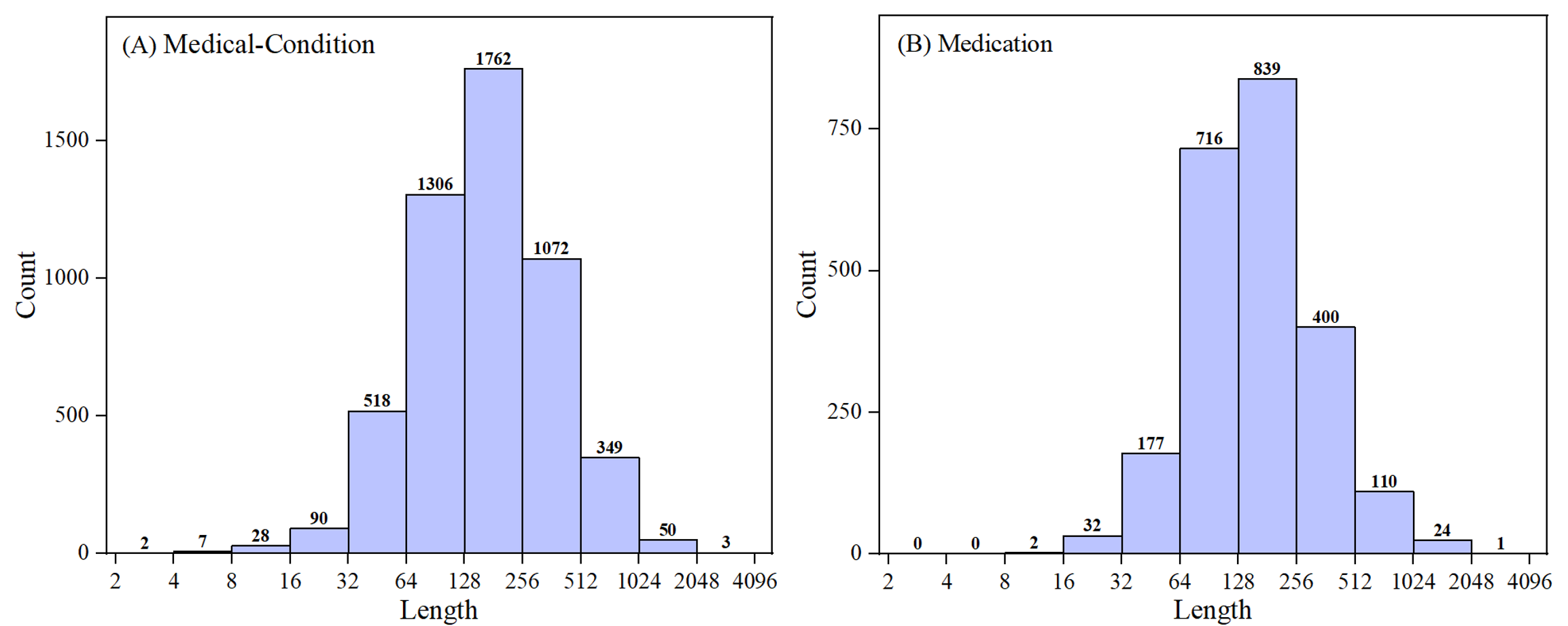}
  \begin{figure}
  \caption{Distribution of sentence length in (A)\textbf{Medical-Condition} and (B)\textbf{Medication} datasets. 
  The Length refers to the number of tokens in sequence generated by the standard BERT tokenizer\cite{ref1}.}\label{fig2}
  \vspace{-1.0em}
\end{figure}
  \end{center}

The text of the datasets is tokenized by BERT tokenizer\cite{ref1}, and the
length statistics of the token sequences are shown in Figure \ref{fig2}. The
original datasets were not divided, so 30\% of each dataset is randomly
selected as the evaluate set. The distribution of the datasets after
division is shown in Table \ref{table3}. All texts in the dataset are intercepted
to 1024 tokens.

\begin{center}
  \newcolumntype{Y}{>{\centering\arraybackslash}X}
  \begin{table}[!htbp]
  \centering
  \caption{Statistics of labels in both datasets.}\label{table3}

  \begin{tabular}{c<{\centering}|c|cc|c}
    \hlineB{2}
    \textbf{Dataset} & \textbf{Label} & \textbf{Train} & \textbf{Test} & \textbf{Total} \\
    \hline
    \multirow{4}{*}{\textbf{Medical-Condition}} & Exist & 1662  & 734   & 2396 \\
                                          & Recover & 501   & 202   & 703 \\
                                          & Deteriorate & 1468  & 621   & 2089 \\
    \cline{2-5}                           & Total & 3631  & 1557  & 5188 \\
    \hlineB{1}
    \multirow{4}{*}{\textbf{Medication}} & Effective & 337   & 125   & 462 \\
                                  & Ineffective & 436   & 117   & 553 \\
                                  & SAE   & 837   & 389   & 1226 \\
    \cline{2-5}                   & Total & 1610  & 631   & 2241 \\
    \hlineB{2}
    \end{tabular}%
    \vspace{-0.5em}
\end{table}%
\end{center}

\subsection{4.2 Regularization}

\textbf{L2 Normalization}\cite{ref11} is one of the most common methods to prevent
overfitting of deep neural networks. L2
normalization is to add the L2 norm of all the weight to the loss
function as a penalty term. If the model parameters are recorded as $\theta$, the loss function with L2 normalization is shown
below, where $r_{L2}$ is the normalization rate of L2, the
hyperparameter of the training.

\begin{equation}
Loss_{L2} = Loss + r_{L2}\sum_{\theta}^{}\| \theta \|_{L2}\\
\end{equation}

\textbf{Dropout} layer\cite{ref12} randomly discards the input
tensor value with a probability $p_{dropout}$ during the
training. The dropout layer has been proved to effect as data augmentation\cite{dropout2}, which improves
the robustness of neural networks, so it is widely used in deep
learning.

\textbf{Label Smoothing} method\cite{ref13} is an overfitting
prevention method for classification tasks. It replaces $0$ in the
label one-hot vector to $\gamma$, and $1$ to $1 - \gamma$ so that
the input of $softmax$ is much smoother to prevent overfitting
for some unbalanced tags.

\subsection{4.3 Experiment Settings}

\begin{center}
  \begin{table}[htbp]
    \centering
    \caption{Model and training hyperparameters.}\label{table4}
  
      \begin{tabular}{p{13.39em}cc}
      \toprule
      \textbf{Hyperparameters} & \textbf{Expression} & \textbf{Value} \\
      \midrule
      Global embedding dimension & $d_{g}$ & $128$  \\
      Local embedding dimension & $d_{l}$ & $768$ \\
      Number of labels & $U$ & $3$ \\
      L2 regularization rate & $r_{L2}$ &  $1\times 10^{-5}$ \\
      Dropout probability & $p_{dropout}$  & $0.3$ \\
      Segment length &  $l$  & $128$ \\
      Segments number & $n$ & $8$ \\
      Total input length & $l\times n$  & $1024$ \\
      Label smoothing factor &  $\gamma$ & $0.2$ \\
      Skimming learning rate & $\alpha_1$ & $1\times 10^{-4}$ \\
      Intensive learning rate & $\alpha_2$ & $1\times 10^{-5}$ \\
      First moment factor & $\beta_1$ & $0.9$ \\
      Second moment factor & $\beta_2$ & $0.99$ \\
      Minibatch size & $n_b$ & $32$ \\
      \bottomrule
      \end{tabular}%
    \vspace{-0.5em}
  \end{table}%
  \end{center}

\textbf{Training of PrC-SaA}. First, the unique token {[}CLS{]} and
{[}SEP{]} of BERT, which is used to declare the classification task and
the end of a sentence, are attached to the beginning and
end of the input text respectively. BERT-Tiny loads from Google
pre-trained model. We use all three regularization methods above to
reduce overfitting. Parameters were updated using Adam optimizer\cite{ref14}, which introduces three model hyperparameters: learning rate $\alpha_{1}$, momentum factor $\beta_{1},\beta_{2}$.

\textbf{Training of SkIn}. After the previous training is finished, the
trained PrC-SaA model is used to distill all the texts in datasets
according to Algorithm \ref{a1}. {[}CLS{]} and {[}SEP{]} are respectively
added to the head and tail of the selected key segment, which are input
into the KSC model, so that the whole SkIn model is jointly trained.
BERT-Base loads from Google pre-trained model. We use all three
regularization methods mentioned above to reduce overfitting too.
Parameters were updated using Adam optimizer too, and the learning rate
is adjusted to $\alpha_{2}$.

Table \ref{table4} shows all the model and training hyperparameters.

At the same time, the time \& memory cost of SkIn under variable input
length is measured. In the measurement of cost, SkIn model adopts two methods
to adapt to the input length: (1) \textbf{SkIn-InvariableSegment} keeps the total number $n$ of segments
fixed, and adjust the length $l$ of each segment in proportion to
the increase of input length; (2) \textbf{SkIn-VariableSegment} keeps the length $l$ of each segment
unchanged, and adjust the total number $n$ of segments in proportion
to  the input length.

\subsection{4.4 Baselines}

The experimental result of SkIn is compared with the following baseline models.

\begin{figure}
\begin{center}
\includegraphics[width=0.95\textwidth]{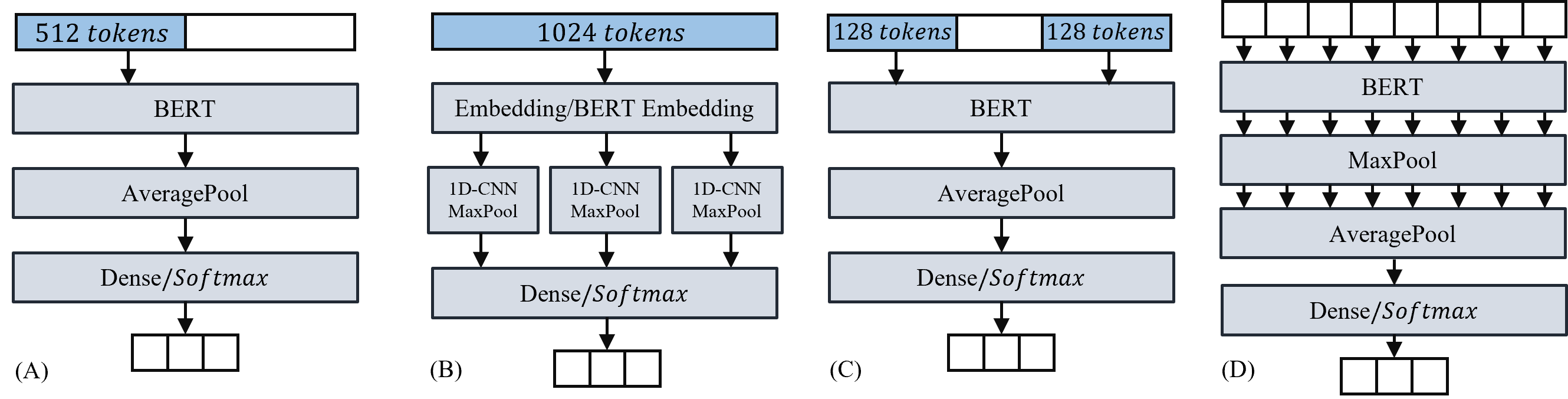}

\caption{Baseline models.\\ (A)BERT (B)CNN/CNN-BERTemb (C)BERT-128HT (D)BERT-SlideWindow}\label{fig3}
\end{center}
\end{figure}

\textbf{BERT}\cite{howto} is shown in Figure \ref{fig3}-(A). The length of all texts is
truncated or padded to the BERT's upper limit of 512 tokens. After
contextualized embedding using the pre-trained BERT-Base model, the
global average pooling is performed, and the pooling result is output
through a dense layer activated by $softmax$.

\textbf{CNN}\cite{ref3} is shown in Figure \ref{fig3}-(B). The input tokens are embedded to
300-dimensional vectors and input into three 1D-CNNs activated by
\textbf{Re}ctified \textbf{L}inear \textbf{U}nit (ReLU)\cite{ref15} with convolution kernel widths of 3, 4 and 5. Then, the
convolution results are output to the global maximum pooling, and the
pooling results are concatenated into the final feature vectors and
input into a dense layer activated by $softmax$ for output.

\textbf{CNN-BERTemb} is similar to the CNN model as shown in Figure \ref{fig3}-(B).
The embedding layer of CNN-BERTemb uses the embedding layer of the
BERT-Base pre-trained model to generate 768-dimensional non-contextual
word embedding vectors. The structure of the other parts of CNN-BERTemb is consistent with CNN model.

\textbf{BERT-128HT}\cite{ref5} is shown in Figure \ref{fig3}-(C). Connect the 128 tokens at
the beginning and 128 tokens at the end of the text with BERT's unique
token {[}SEP{]} (if the length is less than 128, pad it to 128). Input
the connected sequence to BERT-Base for contextualized embedding,
then perform global average pooling, and output the result through a
dense layer activated by $softmax$.

\textbf{BERT-SlideWindow}\cite{howto} is shown in Figure \ref{fig3}-(D). The length of each
text is truncated or padded to 1024 tokens and then divided into 8
segments with equal lengths of 128. Each segment is embedded in BERT-Base
and input into a global maximum pooling layer. After that, these
pooling results are averaged, and the average result outputs through a
dense layer activated by $softmax$.

\section{5. Result}

\subsection{5.1 Result of PrC-SaA}

The classification accuracy of PrC-SaA is shown in Table \ref{table6}. The accuracy of PrC-SaA is not satisfactory, 
but we pay more attention to whether it can select the key segment correctly rather than the classification accuracy within the PrC-SaA stage.

Select a piece of text in the Medical-Condition test set, input it into
PrC-SaA model, and make it generate scores for different segments of the
text, as shown in Table \ref{table5}. The true label of this text is \textbf{Deteriorate},
and the classification predicted by the PrC-SaA model is also
\textbf{Deteriorate}. 

As underlined in Table \ref{table5}, the author emphasizes
the deterioration of his symptoms in the second segment, so the PrC-SaA
model gives it the highest score. This example can prove the
effectiveness of the PrC-SaA in selecting the key segment from texts.

\begin{center}
  \begin{table}[!htbp]
  \vspace{-0em}
  \caption{Example of segmentation performance of PrC-SaA. \\ The {[}Pad{]} is the padding token used to pad the length of the input sequence. The underlined parts are the key sentences judged manually.}\label{table5}
  \vspace{-2em}
\end{table}%
\includegraphics[width=\textwidth]{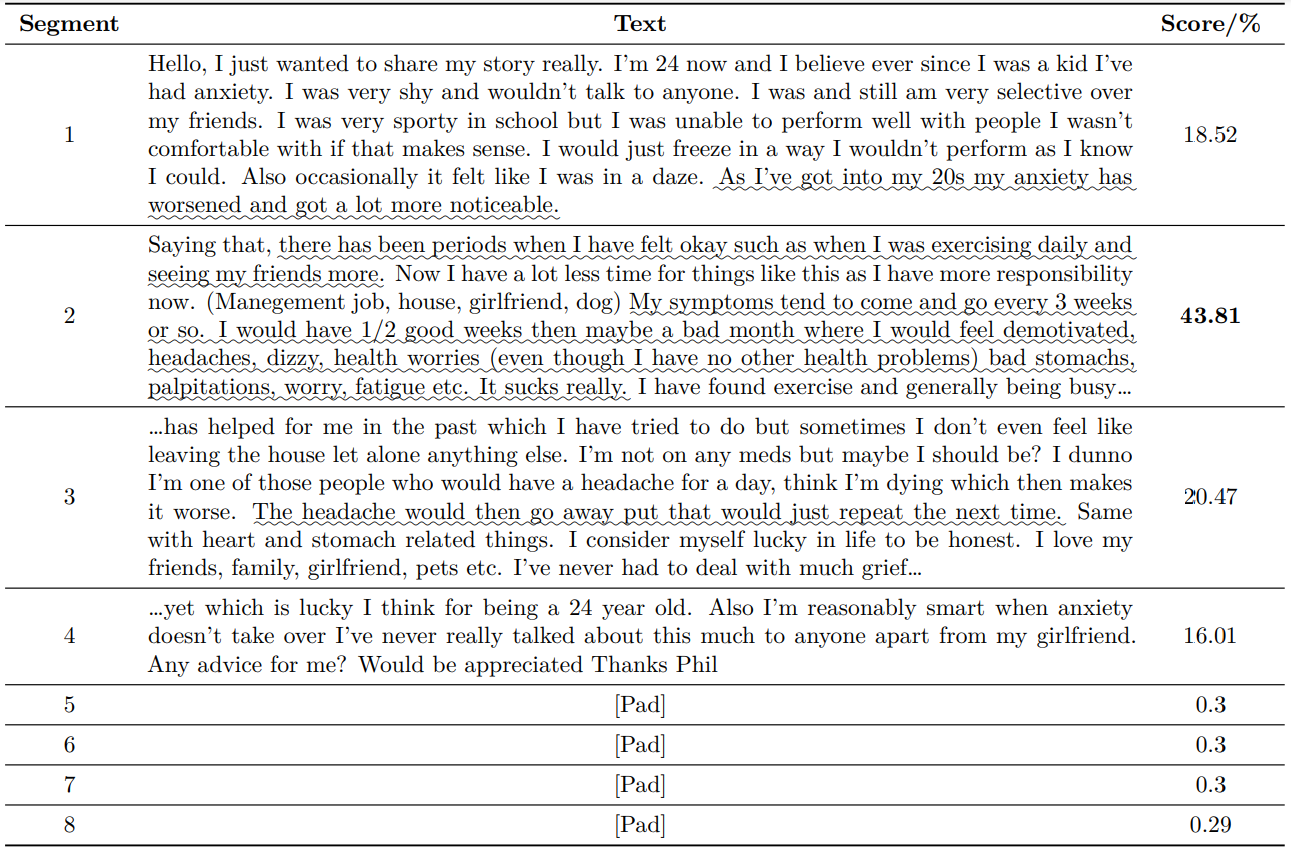}
\end{center}

\subsection{5.2 Result of SkIn}\label{seg52}

The classification performance scores of SkIn compared with the baseline
model are shown in Table \ref{table6}.

The results show that the SkIn model significantly improved in both datasets compared to the baseline model. Compared with the CNN model,
the accuracy of the two datasets has improved by about 4\% - 7\%, and
the F1 score about 5\% - 8\%. Compared with the PrC-SaA part, the results
of joint training have been improved by 3\% - 5\% for both accuracy and F1 score, which shows the
effectiveness of the intensive part of SkIn method.

\begin{center}
  \begin{table}[!htbp]
    \vspace{-0em}
    \caption{Classification performance of baselines and our models.\\The best results are shown in bold.}\label{table6}
    \centering
      \begin{tabular}{c|c|c|c|c}
      \hlineB{2}
      \multirow{2}[0]{*}{\textbf{Model}} & \multicolumn{2}{c|}{\textbf{Medical-Condition}} & \multicolumn{2}{c}{\textbf{Medication}} \\
      \cline{2-5}     \multicolumn{1}{c|}{} & \multicolumn{1}{c}{\textbf{Acc./\%}} & \multicolumn{1}{c|}{\textbf{F1/\%}} & \multicolumn{1}{c}{\textbf{Acc./\%}} & \multicolumn{1}{c}{\textbf{F1/\%}} \\
      \hlineB{1}
      BERT  & \multicolumn{1}{c}{61.59} & \multicolumn{1}{c|}{58.02} & \multicolumn{1}{c}{75.54} & \multicolumn{1}{c}{69.99} \\
      CNN   & \multicolumn{1}{c}{57.29} & \multicolumn{1}{c|}{53.04} & \multicolumn{1}{c}{75.69} & \multicolumn{1}{c}{70.18} \\
      CNN-BERTemb & \multicolumn{1}{c}{61.72} & \multicolumn{1}{c|}{55.78} & \multicolumn{1}{c}{77.56} & \multicolumn{1}{c}{72.64}\\
      BERT-128HT       & \multicolumn{1}{c}{63.01} & \multicolumn{1}{c|}{60.09} & \multicolumn{1}{c}{76.41} & \multicolumn{1}{c}{72.17} \\
      BERT-SlideWindow & \multicolumn{1}{c}{60.89} & \multicolumn{1}{c|}{56.57} & \multicolumn{1}{c}{77.42} & \multicolumn{1}{c}{71.91} \\
      \hlineB{1}
      PrC-SaA          & \multicolumn{1}{c}{61.27} & \multicolumn{1}{c|}{56.26} & \multicolumn{1}{c}{76.99} & \multicolumn{1}{c}{72.70} \\
      \textbf{SkIn} & \multicolumn{1}{c}{\textbf{64.42}} & \multicolumn{1}{c|}{\textbf{61.60}} & \multicolumn{1}{c}{\textbf{79.45}} & \multicolumn{1}{c}{\textbf{75.74}} \\
      \hlineB{2}
      \end{tabular}%
      \vspace{-1em}
  \end{table}%
  \end{center}

\subsection{5.3 Time \& Memory Cost}\label{seg53}

We measure the time \& memory cost of BERT, BERT-SlideWindow, and
SkIn with both above-mentioned length adjustment methods on the Medical-Condition
train set. With the increase of token length, similar to the SkIn-InvariableSegment, 
the BERT-SlideWindow model always divides the input into 8
segments, and the length of each segment increases in equal proportion.
Due to the limitation of the algorithm and GPU memory, BERT cannot be trained when
the token length is greater than 512, BERT-SlideWindow runs out of the GPU
memory when the token length is greater than 1536, and
SkIn-InvariableSegment reaches the maximum length of BERT when the
length is greater than 2560. Therefore, the existing data are used for
quadratic polynomial regression to predict the data when the model fails.
The data is presented in Fig. \ref{fig4}.

As shown in Fig.\ref{fig4}, results (A) and (B) show that when the token length is from 128 to 2048, the
cost of both BERT and BERT-SlideWindow show noticeable quadratic growth,
and the cost of SkIn method in this range is much lower than the other
two. Especially with a longer token length, SkIn saves up to more than 90\% (from 162.0 GB to 14.03 GB) of
the memory compared with BERT. Therefore, SkIn can significantly
reduce the cost of BERT pre-trained model in long-text classification.

Results (C) and (D) show that if the number of segments is adjusted to fit the input while the length of each
segment is unchanged, the SkIn method can reach a linear space-time complexity, 
but the division of segments will introduce additional time
cost. Moreover, by changing the length of segments while the number of segments
is maintained, the SkIn method still has a quadratic space-time complexity, 
but does not introduce additional time cost. Therefore, in
actual use, the number of segments and the length of segments should be
adjusted simultaneously according to the text length.

\section{6. Conclusion and Discussion}

Because of the benefits of opinion mining in medical comments mentioned before, we try to use the pre-trained BERT to accomplish the task.
However, the comments in the medical corpus may be lengthy. BERT, which is quadratic to the sequence length, is hard to applicate directly to the long text. So the main contribution of this paper is as follows: 

1. \textbf{The self-adaptive attention (SaA) mechanism} is proposed to select the most critical segment by the classification process without additional dataset annotations.
Based on the SaA and BERT, the Skimming-Intensive (SkIn) method for long text classification is proposed.

2. \textbf{Classification accuracy.} The classification accuracy of SkIn is evaluated on the CMS dataset\cite{ref3}.
The result shows that compared with the baselines, the accuracy of SkIn is improved. 

\begin{center}
  \includegraphics[width=0.85\textwidth]{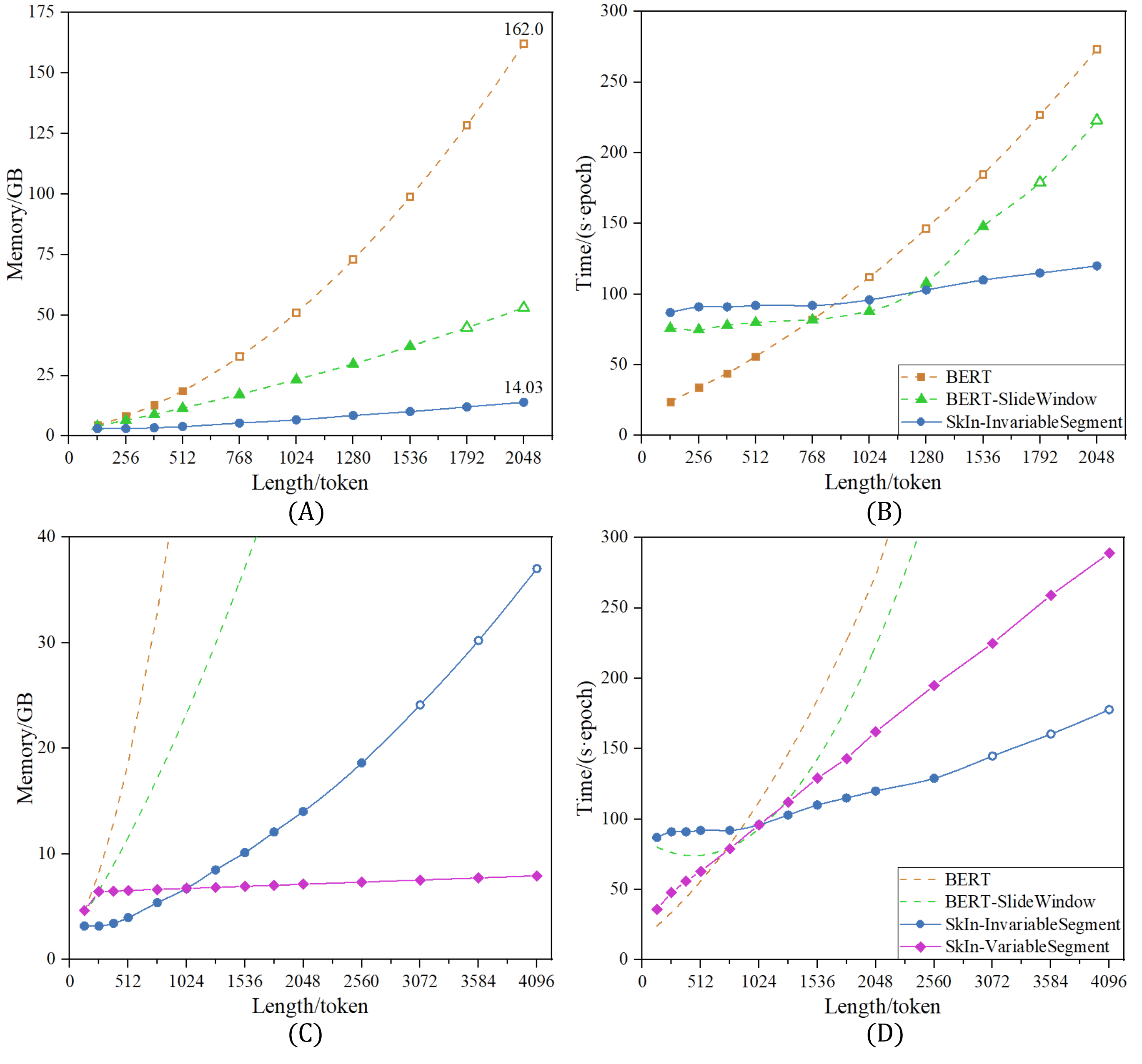}

  \begin{figure}
      \caption{(A)(C) memory and (B)(D) time cost with varying sentence length. The Length refers to the number of tokens in sequence generated by the
      BERT tokenizer. Hollow points represent the data predicted by quadratic polynomial
      regression. The data were measured using 1 NVIDIA Ampere A100 GPU (39GB)
      with batch size of 16.}\label{fig4}
  \end{figure}
  \end{center}

3. \textbf{Cost.} SkIn and baselines methods are compared in terms of space-time cost. The result shows that SkIn saves considerable time and memory
and reaches a linear time-space complexity.

The experiment results show that the SkIn method not only achieves better accuracy than the baseline model but also has 
a linear increase in space-time complexity on long-text datasets, which solves the overflow problem of basic BERT on long texts. 
The self-adaptive attention mechanism can effectively alleviate the problem of lengthy text processing laconically without extra annotations.
This method that only uses an additional network to select the critical segment can be adapted to various NLP downstream models and may become a general paradigm for lengthy text processing.
In future work, we will apply this methodology to complex downstream tasks such as aspect-based sentiment analysis and Q\&A system.

\end{document}